# DUTCH NAMED ENTITY RECOGNITION AND DE-IDENTIFICATION METHODS FOR THE HUMAN RESOURCE DOMAIN

Chaïm van Toledo, Friso van Dijk, and Marco Spruit

Utrecht University, Utrecht, the Netherlands

## ABSTRACT

*The human resource (HR) domain contains various types of privacy-sensitive textual data, such as e-mail correspondence and performance appraisal. Doing research on these documents brings several challenges, one of them anonymisation. In this paper, we evaluate the current Dutch text de-identification methods for the HR domain in four steps. First, by updating one of these methods with the latest named entity recognition (NER) models. The result is that the NER model based on the CoNLL 2002 corpus in combination with the BERTje transformer give the best combination for suppressing persons (recall 0.94) and locations (recall 0.82). For suppressing gender, DEDUCE is performing best (recall 0.53). Second NER evaluation is based on both strict de-identification of entities (a person must be suppressed as a person) and third evaluation on a loose sense of de-identification (no matter what how a person is suppressed, as long it is suppressed). In the fourth and last step a new kind of NER dataset is tested for recognising job titles in tezts.*

## 1. INTRODUCTION

De-identification of texts has become a common task within the medical domain, for researching health care records and many other medical documents. The HR field also contains a lot of textual data, but in contrast to de medical domain we did not find any HR text de-identification tools. De-identification of texts provides benefits for organisations. First, the General Data Protection Regulation (GDPR) asks for limiting personal identifiers (PIDs) in processing data. Second, data scientists can work safer with pseudonymised texts, because the PIDs are suppressed. De-identifying methods will reduce the impact of data breaches. And third, organisations can temper employee privacy concerns by removing individual characteristics. Therefore, to realise these benefits, we investigate text de-identification in Dutch governmental HR e-mail correspondence.

This paper transfer existing Dutch medical text de-identification methods to the HR domain. The case organisation is a Dutch governmental HR organisation. The organisation provides employee payments and HR management. More than 100,000 civil servants rely on their systems. Civil servants could contact the organisation's contact centre by phone, e-mail, and chat. The contact centre gets questions like: "how to get tax reduction by buying a bike for commuting?"

The correspondence system saved more than 300,000 e-mails. Over 2,000 e-mails are manually annotated on eleven characteristics. Next, we trained NER models to recognise names, locations, and organisations. Then, different NER methods and de-identification methods are combined and compared with each other.

This paper also offers a new NER dataset, for recognising job titles. In an unsupervised way, a new Dutch NER dataset is created from vacancies. With the dataset, it is possible to train a model for recognising job titles in texts.

The research question of this paper is: To which extent can current de-identification methods anonymise Dutch HR related texts? Our main contributions are bringing de-identification to the HR domain, benchmarking current de-identification methods, updating NER models, and benchmarking NER methods.

The paper is structured as follows: The Background section elaborates on NER, de-identification and policy and PIDs. From this background, the expectation is that updating NER models with the so-called transformers should enhance de-identification results. About PIDs, none of the de-identification methods are specialised in supressing gender, job titles and regular titles. Because of the absence of recognising the classes, we expect that performance will be modest. In the Methods section, this paper elaborates how we update the NER methods and how we combine this with de-identification methods. The Methods section also explains how the Dutch job title recognition dataset is constructed.

The Results section shows the performances of the NER and de-identification methods. The results are divided in four evaluations. The first one evaluates the performance of the current and the updated NER models. The second evaluation shows how the state-of-the-art de-identification methods suppress PIDs correctly. Third evaluation brings the result of to which extend the PIDs are suppressed, no matter how it is labelled by the methods. The fourth evaluation shows the results of the job title recognition model.

After, we discuss the shortcomings of the current de-identification methods and what features future de-identification methods should have. We conclude that current Dutch NER methods can be improved through transformers and that with the TKS method and a state-of-the-art NER method the best results can be achieved in de-identifying HR texts.

## 2. BACKGROUND

This section overviews three elements of de-identification. The background starts with NER. NER recognise privacy sensitive elements in texts. Second part elaborates about the de-identification methods and the usage of NER systems in de-identification methods. Third and last part explain about the policies of de-identification.

### 2.1. NER For DUTCH

A definition of NER is: "the task of automatically identifying names in text and classifying them into a pre-defined set of categories" [1, p. 1]. Automated text de-identification and NER share the same goal: recognise entities in texts [2]. Three main approaches can be distinguished [3]. The first approach is rule-based, this method contains name lists, regular expressions and other predefined handcrafted rules. Second is an unsupervised learning approach, with clustering, word groups can construct based on context similarity. Third approach is a supervised learning, with predefined labels on words in a corpus. The results are good in this approach, but the construction of corpora for NER is a time-consuming process.

For supervised learning NER, the conditional random fields (CRF) algorithm plays a major role. CRF takes neighbouring tokens into account and so context plays a role in labelling names in texts. Long short-term memory (LSTM) also impacts a major role in NER. LSTM is "capable of remembering information over long time periods during the processing of a sequence" [4, p. 1]. Recent developments show an upcoming rise of transformers, like BERT (Bidirectional Encoding Representations for Transformers), introduced by Google [5]. Test results show improvements in NER because of BERT.

Two known hand-annotated corpora are publicly available to create Dutch NER systems, namely CoNLL-2002 (Conll) and SoNaR-1 (Sonar). Conll has four labels for entity recognition: persons, locations, organisations and miscellaneous. The training corpus contains 218,737 lines of words [6]. The Sonar corpus got two extra labels: products and events. Sonar contains 1 million words. Where the Conll corpus uses news data, Sonar uses news items, manuals, autocues, fiction and reports and 'new' media like blogs, forums, chat and SMS [1].

Conll and the Sonar dataset tags their entities in the IOB2 format. Word or tokens are represented vertically, with first the word and second the entity category. O is for outside the scope (or the non-categorised tokens). B-PER (B for begin) stands for name and can be extended with I-PER if the name contains more than one token [7].

For Dutch, there is a well-known NER system, namely FROG [8]. FROG detects persons, organisations, locations, products, events, and miscellaneous entities in texts. Another Dutch NER system (with English, Spanish and German) is created by Lample et al. [9], but unfortunately, the Dutch model isn't available online. Another attempt is Polyglot [10], based on Wikipedia and freebase data and brings a service with 40 language models. Performances for NER are mostly tested with the Conll corpora. Table 1 lists the NER systems with support for Dutch.

Table 1: NER systems with support for Dutch

| Approach | Test set | Language | Rates |
|---|---|---|---|
| Classifying Wikipedia data [11] | CoNLL-2003 | English | $F_1$-score 85.2 |
| | CoNLL-2002 | Dutch | $F_1$-score 78.6 |
| | CoNLL-2003 | German | $F_1$-score 66.5 |
| | CoNLL-2002 | Spanish | $F_1$-score 79.6 |
| Single CRF classifier [1], [8] | SoNaR | Dutch | $F_1$-score 84.91 |
| Discriminative Learning, word embeddings [10] | CoNLL-2003 | English | $F_1$-score 71.3 |
| | CoNLL-2002 | Dutch | $F_1$-score 59.6 |
| | CoNLL-2002 | Spanish | $F_1$-score 63.0 |
| LSTM-based [12] | CoNLL-2003 | English | $F_1$-score 84.57 |
| | CoNLL-2002 | Dutch | $F_1$-score 78.08 |
| | CoNLL-2003 | German | $F_1$-score 72.08 |
| | CoNLL-2002 | Spanish | $F_1$-score 81.83 |
| LSTM-CRF [9] | CoNLL-2003 | English | $F_1$-score 90.94 |
| | CoNLL-2002 | Dutch | $F_1$-score 81.74 |
| | CoNLL-2003 | German | $F_1$-score 78.76 |
| | CoNLL-2002 | Spanish | $F_1$-score 85.75 |

## 2.2. Text De-Identification Methods

Because of the Health Insurance Portability and Accountability Act (HIPAA) regulations, text de-identification methods originate from the medical domain. These methods use a combination of rule-based and/or statistical approaches. Both approaches are used in the method of Tjong Kim Sang (TKS) et al. [13]. In this method, the user can add names in a list and can also find names with FROG NER. There are also non-statistical approaches like DEDUCE [14]. This rule-based method employs user-added lists of names and other sensitive entities. Table 2 gives an overview of the two different approaches in Dutch.

DEDUCE is created for de-identifying psychiatric nursing notes. The method requires an organisational context for defining important entities. The method brings the most popular names in the Netherlands, but it is recommendable to expanse this list with person names from the organisational systems. The same routine implies for institutions or organisations and locations.

TKS uses NER systems, like Frog, for tokenising the text document. The output is directly annotated in long lists of tokens with labels (entity or O). Like the case of DEDUCE, extra organisational context extends the method. The tokens are already annotated by the NER system or will be looked up in the context lists. At last, the tokens will be analysed with regular expressions for dates, (phone) numbers and e-mails.

Table 2: Overview of previous Dutch text anonymisation research.

| Name | Approach | Corpus | Identifiers | Rates |
|---|---|---|---|---|
| DEDUCE [14] | Rule-based | Psychiatric nursing notes (2,000) | Person, url, institution or organisation, location, phone number, age, date | $F_1$-score 0.826 |
| TKS [13] | Rule-based, CRF | Data from therapeutic sessions | Frog: Persons, locations, events, misc., products, organisations Method: numbers, months, days, numbers, month, days, mails, phone numbers | $F_1$-score 0.84 unlabelled score, low micro average 0.55 |

Table 3: HIPAA elements [17]

| Personal identifiable elements | |
|---|---|
| Names | All geographical subdivisions smaller than a State |
| All elements of dates (except year) for dates that are directly related to an individual | Telephone and fax numbers |
| Online identifier (such as e-mail) | Identification number / social security numbers |
| Medical record numbers | Health plan beneficiary numbers |
| Account numbers | Certificate/license numbers |
| Vehicle identifiers and serial numbers | Device identifiers and serial numbers |
| Web Universal Resource Locators | Internet Protocol (IP) addresses |
| Biometric identifiers, including finger and voice prints | Full face photographs and any comparable images |
| Any other unique identifying number, characteristic, or code | |

## 2.3. Policy And PIDs

De-identified data contains no identifiable elements of individuals. When an element is identifiable to an individual differs from situation to situation, and is also referred to as an anonymity versus utility dilemma [15]. The medical world introduced many de-identifiers. It is helpful to see what criteria there are, because the GDPR does not have an exact specification about what PIDs are. The law describes that personal data: "any information relating to an identified or identifiable natural person" [16, p. 2]. The HIPAA provides eighteen possible identifiers, displayed in table 3[17].

De-identified data asks for policy. It would be dangerous to give an anonymised dataset to the public. For example, Narayanan and Shmatikov [18] demonstrate a de-anonymisation attack on the Netflix user dataset. Netflix de-identified user data for a recommender system competition, but the public release of this dataset, backfired to the corporation. Hence, three important policies should be considered when working with de-identified data [19]. First, control spread of the data. Second, prevent identification attempts. Third, provide security measurements.

## 3. METHODS

The methods explain how the evaluation is done. The first two sections explain how the NER is constructed and how the NER and the de-identification methods are evaluated. The fourt part elaborates about the metrics. The third part give an inside in how the job title recognition dataset is constructed. The fourth part explains how the data is annotated and how the agreement is scored between the annotators.

### 3.1. NER Construction and Evaluation

As explained in Background, two Dutch NER applications are publicly available, Polyglot and FROG. We construct four NER models based on the Conll and Sonar corpora. The first two models on both the corpora are trained on the BERT Multilingual Cased (ML), where Dutch is included. The third and fourth models are trained on the two corpora with BERTje [20]. BERTje has specifically been created for the Dutch language. From both Conll and Sonar, we only test persons, locations and organisations. Both contain miscellaneous, but this entity is not annotated in the test dataset. Sonar also contains products and events, but these are also disregarded from the test dataset. Software for constructing NER transformers based on BERT is derived from Raj's github [21].

### 3.2. De-Identification Evaluation

Based on the literature, this paper evaluates two de-identification methods, DEDUCE and the method of TKS. The method of TKS uses Frog, but other NERs can also be attached to this method. Therefore, the created NER models based on BERT, are also attached to the method of TKS.

Due to a lack of comparability between DEDUCE and TKS, this research chose not to evaluate DEDUCE's url, phone number and age retrieval performance. DEDUCE can detect somebody's age, or phone number, but there are also other forms of numbers. DEDUCE classifies e-mail and websites as URL, therefore it is difficult to distinguish them from each other. With the method of TKS, this research did not evaluate phone numbers, mails, events, miscellaneous and products, for the same reasons as with DEDUCE. In table 4, our annotation classification labels connects to the DEDUCE and TKS label.

Table 4: Comparison of entity classification identifiers.

| Our classification labels | DEDUCE | TKS |
|---|---|---|
| Person | Person | Person |
| Organisations | Institution | Organisation |
| Location | Location | Location |
| Num | ... | Num |
| Date | Date | Month, date, days numbers |

### 3.3. Job Title Recognition Dataset

A new dataset for Dutch NER is created for recognising job titles in texts. This is done in four steps. First by collecting Dutch governmental vacancies. Second step is creating a list of job titles from the Dutch governmental vacancies. In more detail, by creating the job title list the job titles with senior in front are copied, without the term senior in front.Then, all job titles without senior in front are copied with senior in front. Then, the job title list got extra job titles, like minister or judge (these job titles are not in the structured job title list, because the job application in this category is not regulary via the website. In total a list of 39,338 job titles are extracted in an order of a descending number of characters.

Third is finding the job titles in the vacancy texts with the list of job titles. In an unsupervised way, the job titles are checked in the vacancy texts. The next sub step tokenised the vacancy texts and give for every token an O or a B- or I-JOBTITLE tag.

The last and fourth step is training a NER model for recognising the job titles in texts. The BERT Base Multilingual cased model was the base for training the model. A F1-score was given of 0.91. The dataset is available for download.

### 3.4. Evaluation Metrics

The test performance evaluation metrics are precision, recall and f1-score. Precision is measured by ($\Sigma$ true positive) / ($\Sigma$ true positive + $\Sigma$ false positive). Recall is measured by ($\Sigma$ true positive) / ($\Sigma$ true positive + $\Sigma$ false negative). F1-score is then measured by $2*((precision*recall)/(precision+recall))$. This paper appoints recall as the most important metric, because the recall is measured with the false negatives, so the harm is higher when there occurs a false negative than when there occurs a false positive. F1-score is the second-most important metric, because of the combination of false negatives and false positives in the measurements.

Table 5: Generated examples of e-mail questions. Signatures are parsed out of the context

| |
|---|
| Goodmorning, Please handle the following. [SIGNATURE] Dear colleague, from the 9$^{th}$ of October 2016, I am seconded to the Apeldoorn office of the Tax and Customs Administration, but in the P-Portal my old Ministry of Finance e-mail (j.doe@minfin.nl) is still connected to my account. I would like my new e-mail (john.doe@belastingdienst.nl) to be connected to P-Direct. My employee number is 98706540. Thanks in advantage. Greetings John. [SIGNATURE] |
| Hello, with permission from the board, I want to change my commuting fee. During long road construction, my commuting distance to the PI Amsterdam increased from 35 km to 41 km. This situation is happening since the 5$^{th}$ of November past year, so I'd like, retrospectively, to get a higher commuting fee since the 5$^{th}$ of November. Sincerely, [SIGNATURE] |

### 3.5. Annotated Data

E-mail data is used from a Dutch government organisation. This e-mail data concerns all kinds of HR related issues. Most of these correspondences relate to HR administration or to modification

to personnel files. The dataset contains 2,017 e-mails, with a mean of 133 tokens per e-mail. Table 5 shows two examples of messages. Annotations were made in these e-mails, in total 13,496. Software for making annotations is AnnotatorJS and a SQL database to store these annotations. There is a mean of 6.69 annotations per e-mail, with a standard deviation of 11.21.

Table 6 contains the different PIDs with the occurrences in the dataset. Date is everything what points to a date, i.e. month year, or day. Weekdays, like Sunday and Monday, are excluded; this turned out too generic. Numbers are defined as a sequence of two or more digits, meaning that 1 or 2 are excluded, but 10 and 222 are not excluded. One digit turned out to be too generic. Persons (per) include names, surnames, and initials. This research also includes Gender, because gender related words can be seen as a binary. With these words and other identifiers an attacker can easily deduce people. Gender words are like sir, madam (or in Dutch: de heer, mevrouw), but also he, she, son, his. We define Organisation (org) as groups of people larger than one person. The E-mail (mail) section was difficult, because some e-mails contain a whitespace, this is strictly impossible, probably written by mistake. Location (loc) is everything what refers to a physical place, such as a street, postal code, municipal, country, area, region and so forth. Job title refers to somebody's job, like policy officer or contact center agent. Title includes degrees such as MSc, Dr or Duke. Code is anything what refers to an account, like IBAN numbers, usernames and passwords. Websites can also reveal the organisation a person is working at, so this research annotates this information as well.

Table 6: Counted PIDs in the dataset

| PID | Date | Num | Per | Gender | Org | Mail | Loc | Job title | Code | Title | Website |
|-----|------|-----|-----|--------|-----|------|-----|-----------|------|-------|---------|
| Num | 3,628 | 3,338 | 3,086 | 1,567 | 1,011 | 279 | 225 | 120 | 116 | 96 | 28 |

The first annotator labelled the entire sample of 2,017 e-mails. A second annotator labelled 283 e-mails, 14% of the sample. For measuring the correctness of the labelled representations, an interrater reliability is used with a kappa statistic. According to [22], a kappa score of at least 0.80 is sufficient, a kappa below 0.60 indicates a non-agreement between annotators. A kappa statistic is measured as follow: $\kappa = \frac{\Pr(a) - \Pr(e)}{1 - \Pr(e)}$

Pr(a) is the observed agreement between the annotators and Pr(e) is the expected change agreement. The observed agreement is 0.99 and the expected agreement is 0.82. The kappa score is 0.92 and we conclude there is a high agreement between the annotators.

## 4. RESULTS

The results are distinguished in four evaluations. First, how well do the NER models perform? Second, how do the de-identification methods perform in a "strict" sense? A strict sense means that the suppression both identifies and classifies the entity correctly. The third evaluation shows how the de-identification methods perform in a "loose" sense. A loose sense only takes the identification of the suppression into account and not the classification. The fourth and last evaluation provides the evaluation of job title recognition.

The first three evaluations are equal important, because a good NER will recognise entities without organisational knowledge. A high strict de-identification ensures that the utility will not drop too much, because only the sensitive entities are suppressed, and the false positives are low as possible. A loose sense de-identification evaluation is important because suppression is the key to anonymisation.

## 4.1. NER Evaluation

Table 8 shows the results of two existing NER models, namely Polyglot and FROG, against trained BERT-based models. A remarkable outcome is the fact that the Conll corpus gives highest results with respect to recall. As the creators of Sonar explained, Sonar has a more diverse corpus than Conll. Because of this diversity, the expectation was that Sonar could better recognise entities than Conll. The precision of Sonar BERT ML gives in almost all cases the best results.

Table 7: NER results (underline is highest result in row)

|  | Metrics | Polyglot | Frog | Conll BERT ML | Sonar BERT ML | Sonar BERTje | ConllBERTje |
|---|---|---|---|---|---|---|---|
| Per | Precision | 0.69 | 0.68 | 0.69 | 0.81 | 0.77 | 0.72 |
|  | Recall | 0.35 | 0.80 | 0.86 | 0.86 | 0.83 | 0.88 |
|  | F1 | 0.46 | 0.73 | 0.77 | 0.83 | 0.80 | 0.80 |
| Org | Precision | 0.66 | 0.38 | 0.47 | 0.71 | 0.63 | 0.59 |
|  | Recall | 0.16 | 0.46 | 0.66 | 0.65 | 0.54 | 0.69 |
|  | F1 | 0.26 | 0.42 | 0.55 | 0.68 | 0.58 | 0.64 |
| Loc | Precision | 0.10 | 0.09 | 0.26 | 0.40 | 0.34 | 0.41 |
|  | Recall | 0.36 | 0.49 | 0.64 | 0.60 | 0.57 | 0.65 |
|  | F1 | 0.16 | 0.15 | 0.37 | 0.48 | 0.42 | 0.50 |

## 4.2. Strict De-Identification

Table 8 shows the results for strict de-identification. This means that for example a person is identified as an organisation, the person identifier gets a false negative. The BERT-based transformers perform better in almost all NER related fields. It is in the line of expectations that the method of TKS would perform the best regarding the recall in combination with the ConllBERTje model. The NER results section shows that this model also performs best.

The precision can be lower compared to the NER results. The reason for this decrease is the added organisational data. This added data has a positive influence on the recall, because more entities are recognised, but the downside is an over fitting and that influences the precision. Or the problem in other words: sometimes names can be regular words.

Table 8: Strict de-identification results (underline is highest result in row)

|  | Metrics | DED UCE | TKS + Frog | TKS + Sonar BERT ML | TKS + Conll BERT ML | TKS + Sonar BERTje | TKS + ConllBERTje |
|---|---|---|---|---|---|---|---|
| Per | Precision | 0.42 | 0.66 | 0.64 | 0.57 | 0.61 | 0.59 |
|  | Recall | 0.77 | 0.86 | 0.91 | 0.87 | 0.88 | 0.92 |
|  | F1 | 0.55 | 0.74 | 0.75 | 0.69 | 0.72 | 0.72 |
| Org | Precision | 0.06 | 0.28 | 0.56 | 0.25 | 0.51 | 0.47 |
|  | Recall | 0.00 | 0.45 | 0.68 | 0.49 | 0.59 | 0.68 |
|  | F1 | 0.00 | 0.35 | 0.61 | 0.33 | 0.55 | 0.56 |
| Loc | Precision | 0.70 | 0.16 | 0.23 | 0.20 | 0.21 | 0.26 |
|  | Recall | 0.26 | 0.38 | 0.32 | 0.47 | 0.31 | 0.40 |
|  | F1 | 0.38 | 0.22 | 0.26 | 0.28 | 0.25 | 0.31 |
| Date | Precision | 0.71 | 0.97 | 0.98 | 0.98 | 0.97 | 0.97 |
|  | Recall | 0.65 | 0.90 | 0.94 | 0.90 | 0.94 | 0.94 |
|  | F1 | 0.68 | 0.96 | 0.96 | 0.94 | 0.96 | 0.96 |
| Num | Precision | ... | 0.62 | 0.62 | 0.62 | 0.62 | 0.62 |
|  | Recall | ... | 0.92 | 0.93 | 0.91 | 0.93 | 0.93 |
|  | F1 | ... | 0.74 | 0.74 | 0.74 | 0.74 | 0.74 |

Table 9: De-identification results (underline is highest result in row)

| Entities | Metrics | DEDUCE | TKS + Frog | TKS + Sonar BERT ML | TKS + Conll BERT ML | TKS + Sonar BERTje | TKS + ConllBERTje |
|---|---|---|---|---|---|---|---|
| Person | Recall | 0.79 | 0.9 | 0.93 | 0.89 | 0.93 | 0.94 |
| Org | Recall | 0.00 | 0.75 | 0.81 | 0.76 | 0.77 | 0.78 |
| Location | Recall | 0.38 | 0.76 | 0.75 | 0.77 | 0.76 | 0.82 |
| Date | Recall | 0.61 | 0.92 | 0.96 | 0.92 | 0.97 | 0.97 |
| Number | Recall | 0.30 | 0.88 | 0.89 | 0.87 | 0.88 | 0.89 |
| Gender | Recall | 0.49 | 0.11 | 0.10 | 0.16 | 0.10 | 0.16 |
| E-mail | Recall | 0.53 | 0.44 | 0.40 | 0.40 | 0.49 | 0.48 |
| Code | Recall | 0.19 | 0.48 | 0.40 | 0.48 | 0.49 | 0.60 |
| Title | Recall | 0.83 | 0.13 | 0.14 | 0.13 | 0.15 | 0.18 |
| Job title | Recall | 0.11 | 0.52 | 0.49 | 0.45 | 0.49 | 0.51 |
| Website | Recall | 0.08 | 0.08 | 0.08 | 0.08 | 0.00 | 0.00 |

## 4.3. Loose Sense De-Identification

The focus in this section is only whether something is suppressed or not. Loose sense de-identification is when an identifier is, suppressed, a true positive is seen and when an identifier is not suppressed, a false negative is seen. False positives are not considered, because not all methods recognise gender, codes, titles and job titles. A false positive can also be a true positive in another identifier class.

The highest results of table 10 are TKS with the Conll corpus and BERTje for training. The combination preforms best on persons and locations. For persons the BERT based methods are preforming slightly better than Frog. The same is happening with locations and organisations.

Interesting is DEDUCE, as it accounts for gender and title more often. This happens because DEDUCE looks for so-called prefixes at names. These prefixes are like sir, prof, dr and so forth. DEDUCE is not build for recognising words as her/his or son/daughter, so the results remain at a recall of 0.49. However, for recognising titles, it does a good job, with a recall of 0.83. The prefix list was incomplete and should be enlarged with more titles (like nobiliary and accountancy titles).

E-mails are not well recognised; this is due to the tokenisation of the NER methods. The example sentence: "My e-mail is johndoe@example.com" will be tokenised as follows: "My", "e-mail", "is", "johndoe", "@", "uu.nl". The TKS method wants a full e-mail as token and simply selects the e-mail based on its @-sign. The DEDUCE method also contains errors for particularly e-mails, for example when somebody writes their e-mail in capitals or partly in capitals.

## 4.4. Job Title Recognition

For the job title recognition task, only one model is trained, Job title recognition in combination with BERT Multilingual. Table 10 give an overview of the outcome with the new model tested on the annotated dataset. The recall is excellent, but the precision is poor. The phenomenon is explainable if the top five most common outcome entities are covering 78% of the prediction output. Words like employee, manager, adviser and human resource supporter. These words are indeed job titles, however, we did not mark them as privacy sensitive, because the words were too generic and not pointing to a person direct.

Table 10: Job title recognition results

| Metric | Job title + BERT ML |
|--------|---------------------|
| Precision | 0.08 |
| Recall | 1 |
| F1 | 0.16 |

## 5. DISCUSSION

De-identification is a difficult task, as which entities are revealing an individual, and which are not, differs per situation. This research aimed to be as strict as possible in annotating identifiers. Unlike other studies, we also took job titles, titles and gender into account. Of course, it is arguable that not every identifier is as important as the other. Revealing a person's name holds more privacy risk than revealing a person's gender. But a combination of generic identifiers can lead to a high enough specificity that allows for the identification of individuals

None of the methods had a hundred percent score compared to the annotated set and this means that using these methods will not de-identify everybody. Hence, when using a selected method, a researcher should always check data for false negatives. Besides the de-identification techniques, it is arguable to embrace text de-identification workflows in a research organisation. With handles to identify which PIDs are important to suppress.

The BERT transformers had in almost all cases a positive influence on enhancing the identification of the person, organisation, and location entities. A particularly interesting potential future improvement in NER method performance could be to focus more on recognising gender in texts. NER models must in the future handle lower cased names because texts are and will always be noisy and we cannot rely that all writing persons will write correctly. Although we underline that updating and expanding NER corpora is a time-consuming process.

We gathered organisational data, like name and organisation lists for feeding DEDUCE and TKS, however, in both methods this turned out insufficient to reach near-perfect performance. For organisational entities the score of DEDUCE was too low. It is arguable that we didn't gather enough organisational data to fill the organisation list. On the other hand, organisational data is never enough. To illustrate, a person can mention his/her partner's name in an e-mail, but a partner's name doesn't have to be saved in the organisation system. So, when requesting all the existing names from the organisation for the names list, a partner's name in an e-mail doesn't have to be matched and thus there can occur a false negative. Hence, in our result, the NER methods could de-identify at least 75 percent, so applying NER always seems at least a good starting point for de-identifying texts.

The major feature of TKS is that it is modular concerning NER applications. We could easily connect BERT-based NER methods to the de-identification method. We recommend that future text de-identification methods should follow this path and consider easy implementations of other entity recognition methods in a pipeline overview. Thereby, it is recommendable to add the possibility to let a user add de-identification methods to the overall de-identification method. The text de-identification task differs per situation.

The job title recognition model gives a good demonstration to suppress all the job titles linking to a person. However, also generic terms like manager, adviser, etc. were suppressed. For doing research this behaviour can do harm to a researcher's dataset. For example analysing performance appraisal, the role of the manager can be interesting. On the other hand, with the model, new information can be extracted for information retrieval.

# 6. CONCLUSIONS

There are no strict rules for what should be left out of a text and what should not. Every word in a text could lead to revealing a natural person. We tried to be strict as possible with our annotations and annotate everything what could lead to identifying a person. However, none of the methods aim to include every PID we identified, like gender or job titles.

The loose sense de-identification section shows that the rule-based approaches can detect a person's gender-based salutation but has more trouble to identify a personal name when it is not in the list. The statistical approaches with NER are way better in the situation when a person's name is not in the provided list.

This paper evaluated Dutch NER models and de-identification methods. None of the methods achieve a near-perfect performance. Updating NER models with transformers had a positive influence in all de-identification approaches. This paper showed that the Dutch Conll corpus gives the best results regarding recall performance in combination with the Dutch pretrained transformer BERTje for de-identifying Dutch HR text data. This paper also contributes a new NER dataset for recognising job titles. The results are looking promising for detecting job titles in texts.

## AUTHORS

**Chaïm van Toledo** is a PhD candidate in the applied question and answering systems of the Department of Information and Computing Sciences at Utrecht University. As a member of the Applied Data Science Lab, he focuses on transforming organisation data to practical datasets for data science and intelligent systems. His main research objective is to enhance and broaden question and answering systems in the Dutch language.

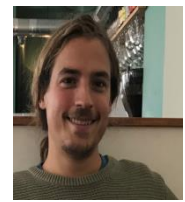

**Friso van Dijk** is a PhD researcher in Privacy Governance at the Department of Information and Computing Sciences at Utrecht University. As a member of the Applied Data Science Lab he works with advanced data science techniques to enhance traditional organizational research methods. His primary research interests are in the development of practical tools for navigating privacy decision-making in the organizational context.

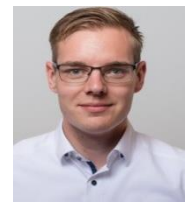

**Dr. Marco Spruit** is an Associate Professor in the Natural Language Processing research group of the Department of Information and Computing Sciences at Utrecht University. As principle investigator in the department's Applied Data Science Lab, his research team primarily works on Self-Service Data Science. Marco's research objective for the coming years is to establish and lead an authoritative national infrastructure for Dutch natural language processing and machine learning to facilitate and popularise self-service data science.

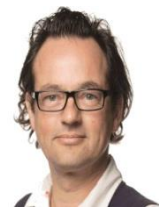